%% file: ijcai26.tex
\documentclass{article}
\usepackage{ijcai26}
\usepackage{bm}
\usepackage{times}
\usepackage{soul}
\usepackage{url}
\usepackage[hidelinks]{hyperref}
\usepackage[utf8]{inputenc}
\usepackage[small]{caption}
\usepackage{graphicx}
\usepackage{amsmath}
\usepackage{amsthm}
\usepackage{booktabs}
\usepackage[table,dvipsnames]{xcolor}
\usepackage{textcomp}
\usepackage{tabularx}
\usepackage{xspace}
\usepackage{enumitem}
\usepackage[ruled,vlined,linesnumbered]{algorithm2e}
\usepackage{cleveref}
\usepackage[switch]{lineno}
\usepackage[referable]{threeparttablex}
\usepackage{multirow}
\usepackage{adjustbox}
\usepackage{makecell}
\usepackage{booktabs}
\usepackage{listings}

\lstdefinestyle{promptstyle}{
  basicstyle=\ttfamily\scriptsize,
  columns=fullflexible,
  breaklines=true,
  breakatwhitespace=true,
  keepspaces=true,
  showstringspaces=false,
  frame=single,
  framerule=0.2pt,
  rulecolor=\color{black!25},
  backgroundcolor=\color{black!2},
  aboveskip=6pt,
  belowskip=6pt
}
\newif\ifcamera
\cameratrue

\ifcamera
\else
  \linenumbers
\fi

\DeclareUnicodeCharacter{2011}{-}

\newcommand{\evolve}[0]{{\texttt{IC3-Evolve}}\xspace}

\definecolor{kwBlock}{HTML}{1F4E79} % deep blue
\definecolor{kwPush}{HTML}{38761D}  % deep green
\definecolor{kwIndG}{HTML}{674EA7}  % deep purple
\definecolor{kwPred}{HTML}{B45F06}  % brown/orange
\newcommand{\algkw}[2]{\textcolor{#1}{\textsc{#2}}}

\title{\evolve: Proof-/Witness-Gated Offline LLM-Driven Heuristic Evolution for IC3 Hardware Model Checking}

\ifcamera
\author{
Mingkai Miao$^{1}$\thanks{Equal contribution.}
\and
Guangyu Hu$^{2}$\footnotemark[1]
\and
Ziyi Yang$^{1}$
\and
Hongce Zhang$^{1,2}$\thanks{Corresponding author.}\\
\affiliations
$^{1}$Hong Kong University of Science and Technology (Guangzhou), Guangzhou, Guangdong, China\\
$^{2}$Hong Kong University of Science and Technology, Clear Water Bay, Hong Kong\\
\emails
\{mmiao815, zyang957\}@connect.hkust-gz.edu.cn,
ghuae@connect.ust.hk,
hongcezh@hkust-gz.edu.cn
}

\else
\author{}
\fi

\begin{document}
\maketitle

\input{sections/abstract.tex}
\input{sections/intro.tex}
\input{sections/preliminary.tex}

\input{sections/method.tex}

\input{sections/experiment.tex}

\input{sections/conclusion.tex}
\input{sections/discussion.tex}

\ifcamera
  \input{sections/ack.tex}
\fi

% \newpage
% \section*{Use of LLM tools}
% We used a large language model to polish the writing and improve clarity. All technical content, experimental results, and claims were verified by the authors, who take full responsibility for the final manuscript.

\bibliographystyle{named}
\bibliography{refs}

\newpage
\section{Appendix}

\nolinenumbers
\subsection{Prompt Templates}
\label{app:prompts}

\subsubsection{Evaluator Agent Prompt}
\begin{lstlisting}[style=promptstyle,caption={Evaluator Agent prompt.},label={lst:evaluator_prompt}]
You are the Evaluator Agent. Given (i) a patch diff and (ii) build/run logs + metrics, decide whether to ACCEPT, REVERT, or RETRY.

Hard semantics / ground rules:
- Treat metrics.runs[*].ok as the pass/fail truth (allowed return codes already reflected there).
- In main.cpp: return code 0 = SAFE, 1 = UNSAFE (not a crash).
- Do not assume the suite is mixed SAFE/UNSAFE; uniform outcomes are valid if consistent with logs/metrics.
- Do not infer expectations from filenames; use only the provided artifacts.

Performance signals:
- Primary: metrics.par2.avg_sec (PAR2 uses 2*timeout for timeouts; lower is better).
- Secondary: solved/timeout counts; optional per-case IC3 stats to explain regressions/improvements.

Decision rubric:
- ACCEPT iff build_ok=true AND all runs ok AND no witness/consistency failures AND performance improves under the regression budget.
- REVERT iff build fails OR any run ok=false OR witness validation fails OR the diff obviously breaks SAFE/UNSAFE semantics.
- RETRY iff results are inconclusive or the patch appears inert (no measurable change and no clear harm).

Output format (STRICT):
- Return ONLY one fenced code block of type json, matching the diagnosis_v1 schema in Listing~\ref{lst:diagnosis_schema}.
- Include: decision, 3-6 numeric reasons, evidence summary, hypothesis evaluation (if provided), and a MoveSet for the next iteration.

MoveSet requirements (brief):
- Provide multiple candidate moves with scores (confidence/risk/cost) and expected metric directions.
- Compass: 2-3 single-slot moves. Jump: >=2 moves from distinct slots.
- Prioritize SLOT_FOCUS if provided; only propose cross-slot moves when justified by evidence.

Inputs (placeholders):
KB: {{KB_SNIPPET}}
DIFF: {{DIFF_SNIPPET}}
LOGS: {{LOGS_SNIPPET}}
METRICS: {{METRICS_JSON}}
BASELINE: {{BASELINE_JSON}}
STATS: {{STATS_JSON}}
HYPOTHESIS: {{HYPOTHESIS_JSON}}
PREV_MOVES: {{MOVES_PROMPT}}
CJ_PLAN: {{CJ_PLAN_SNIPPET}}

\end{lstlisting}

\subsubsection{Programmer Agent Prompt}
\begin{lstlisting}[style=promptstyle,caption={Programmer Agent prompt.},label={lst:programmer_prompt}]
You are the Programmer Agent evolving the IC3 repository.

Goal:
- Propose ONE minimal, safe improvement to IC3 implementation heuristics (or instrumentation) within the selected slot(s).

Hard constraints:
- Small patch: <=80 added lines, <=3 files; must compile with existing Makefile.
- Edit only project sources (.c/.cc/.cpp/.h/.hpp). Do NOT touch: aiger/, minisat/, scorr_aig/, test/.
- Keep SAFE/UNSAFE semantics: in main.cpp, return 0 = SAFE, 1 = UNSAFE.
- Prefer changes within IC3*/Model*/main.cpp unless explicitly justified.

Slot discipline:
- Primary slot must be declared in Hypothesis.json (see Listing~\ref{lst:hypothesis_schema}); optional cross-slot touches must be listed and kept surgical.
- If SLOT_FOCUS is set, stay within it unless strictly necessary.

Instrumentation (optional but encouraged):
- Print one-line counters as: `. HYP <key>: <value>` for parsing.

Required output:
- Edit the repository files directly AND create Hypothesis.json at repo root.
- Do NOT return a diff in the response.

Inputs (placeholders):
REPO MAP: {{CODE_SNIPPET}}
KB: {{KB_SNIPPET}}
LATEST DIAGNOSIS: {{DIAGNOSIS_SNIPPET}}
MOVESET: {{MOVES_PROMPT}}
CJ_PLAN: {{CJ_PLAN_SNIPPET}}
SLOT_FOCUS: {{SLOT_FOCUS}}
\end{lstlisting}

\subsection{Context Management for Prompt Construction}
\label{app:context_management}
LLM-driven code evolution must operate under a finite context window while handling repository-scale code, per-instance logs, and retrieval-augmented notes.
To keep \evolve robust across long runs, we use a hierarchical truncation strategy that assigns an explicit budget to each prompt component and degrades gracefully when overflow is detected.

\paragraph{Hierarchical prompt budgeting.}
We prioritize high-signal artifacts and cap every input by length (line/character) limits:
(i) \emph{Selective code context:} we provide a brief repository summary plus local snippets around the target functions and a small diff excerpt, avoiding full-source injection;
(ii) \emph{Log trimming:} we include detailed logs only for the slowest top-$k\%$ cases and truncate each case by both a maximum number of lines and characters; build/run summaries are also clipped;
(iii) \emph{Slot-aware KB retrieval:} knowledge-base snippets are retrieved by slot relevance (path/slot priority) with a hard length cap (and can be disabled when needed);
(iv) \emph{Structured stats/diffs:} metrics, per-case statistics, and diffs are truncated to essential fields and subject to hard caps.
\Cref{alg:prompt_budget} summarizes the resulting prompt assembly, where \textsf{KB} denotes a slot-specific knowledge base (curated notes/code excerpts/papers per slot) and \textsf{slot focus} specifies the currently editable slot(s).
\input{algorithms/context_window_handle.tex}

\paragraph{Automatic slim-prompt fallback.}
If the framework detects a context overflow (or missing/invalid agent output), it automatically retries with a \emph{slim prompt} that omits KB snippets and aggressively shortens logs/metrics/moveset/changed-file excerpts.
Importantly, this truncation affects only the agent-facing diagnosis and guidance; proof-/witness-gated validation and benchmarking are executed on the full, untruncated artifacts.

\subsection{Extended Case Study: Auto-Evolving Push Propagation}
\label{app:case_study_push}

\noindent\textbf{Setup.}
We study how \evolve improves IC3's push propagation loop when only the \textsc{PushClauses} slot is editable.
Specifically, we zoom into rounds r000--r013 of a representative evolution run on the HWMCC optimization workload (Set~A).
All other IC3 components are frozen, and every candidate is subject to the same proof-/witness-gated validation as in the main paper.

\medskip
\noindent\textbf{Milestones.}
Table~\ref{tab:push_milestones_appendix} summarizes the best-so-far solver variants (green bars in \Cref{fig:evolve_iteration_example}).
Even within a single slot, progress is \emph{punctuated}: many proposals are rejected, while a few accepted edits yield discrete drops in PAR2.

\begin{table}[h]
\centering
\small
\setlength{\tabcolsep}{6pt}
\renewcommand{\arraystretch}{1.05}
\caption{Milestones during the push-propagation phase (Set~A). PAR2 is in seconds (lower is better); TO denotes timeouts.}
\label{tab:push_milestones_appendix}
\begin{tabular}{lcc}
\toprule
\textbf{Round} & \textbf{PAR2 (s)} & \textbf{TO} \\
\midrule
r000 (baseline) & 1050.61 & 25 \\
r002 (Stage 1)  & 1050.42 & 25 \\
r004 (Stage 2)  & 983.73  & 22 \\
r013 (Stage 3)  & 943.07  & 21 \\
\bottomrule
\end{tabular}
\end{table}

\medskip
\noindent\textbf{High-level pattern.}
Across the three accepted edits, the evolution converges on a consistent scheduling principle:
\emph{avoid low-yield cleanup and reallocate effort toward frames that show evidence of progress.}
Notably, the first accepted change yields only a marginal PAR2 improvement (1050.61$\rightarrow$1050.42), but it introduces a measurable signal
that later edits exploit to make more aggressive stall-aware decisions.

\subsection*{Stage 1 (r002): Observe \& Gate Simplification}

\noindent\textbf{Motivation.}
Simplification/cleanup inside propagation can be expensive, and its benefit can be low when the propagation loop is not moving clauses.
A natural first step is to \emph{instrument} progress and gate expensive work on that signal.

\noindent\textbf{Auto-learned rule (simplified).}
{\footnotesize
\begin{verbatim}
if pushedThisRound or
   periodicCheckpoint():
    simplify()
record pushSuccessRate()
\end{verbatim}
}

\noindent\textbf{Why it helps.}
This edit preserves IC3's proof logic (no change to SAT checks or clause soundness), but reduces wasted cleanup on rounds with no movement.
More importantly, it creates a lightweight progress proxy (e.g., push success rate) that makes subsequent stall-aware policies implementable and auditable.

\subsection*{Stage 2 (r004): Skip Repeatedly Stalled Frames}

\noindent\textbf{Motivation.}
Once progress can be observed, the next bottleneck is repeated work on frames that consistently fail to push any clause forward.
Treating repeated failure as a strong negative signal allows the solver to stop spending budget on ``hopeless'' frames \emph{within the current round}.

\noindent\textbf{Auto-learned rule (diff-style, simplified).}
{\footnotesize
\begin{verbatim}
if stallStreak[i] >= limit: continue
pushFrame(i)
updateStallStreak(i)  // reset
                      // else ++
\end{verbatim}
}

\noindent\textbf{Why it helps.}
This change makes the loop explicitly stall-aware: instead of uniform effort across frames, it redirects attempts toward frames that still show movement.
Empirically, this corresponds to the first major jump in Table~\ref{tab:push_milestones_appendix} (TO 25$\rightarrow$22; PAR2 1050.42$\rightarrow$983.73),
consistent with reduced churn in propagation.

\subsection*{Stage 3 (r013): Stall-Aware Adaptive Budgets}

\noindent\textbf{Motivation.}
Stage~2 uses a binary decision (skip vs.\ try). A natural refinement is to turn this into a \emph{graded} allocation policy:
give more attempts to promising frames and cut budgets aggressively for deeply stalled frames, with early termination when marginal returns are low.

\noindent\textbf{Auto-learned rule (diff-style, simplified).}
{\footnotesize
\begin{verbatim}
budget <- adaptBudget(i)
// stall/success history
while clausesRemain(i) and budget>0:
    attemptPush(i)
    budget--
if earlyCut(i): break
\end{verbatim}
}

\noindent\textbf{Why it helps.}
Compared to the skip/try rule, adaptive budgets continuously reallocate effort based on observed yield.
This yields a further (smaller but consistent) improvement (PAR2 983.73$\rightarrow$943.07; TO 22$\rightarrow$21),
suggesting that fine-grained budget control captures additional efficiencies beyond simple stall skipping.

\subsection{What This Reveals About Offline LLM-Driven Code Evolution}

\noindent\textbf{From measurement to control.}
The accepted edits follow a coherent progression: (i) introduce a measurable proxy of progress, (ii) use it to avoid repeated low-yield work,
and (iii) refine the policy into a graded budget allocator. This ``instrument $\rightarrow$ gate $\rightarrow$ adapt'' pattern is typical in human heuristic engineering,
yet here it is discovered through iterative propose--evaluate--promote under strict gates.

\noindent\textbf{Compositionality under constraints.}
All changes are slot-restricted and easy to audit, and they compose cleanly because each stage refines the same underlying scheduling objective
(yield-aware propagation) rather than introducing unrelated tweaks. This helps explain why the framework can accumulate gains without destabilizing the solver.

\noindent\textbf{Takeaway.}
Even when only push propagation is editable, \evolve discovers simple, readable rules that reduce wasted work and reallocate effort toward productive frames,
achieving a measurable reduction in PAR2 and timeouts while leaving IC3's proof logic unchanged.

\end{document}

%% file: sections/abstract.tex
\begin{abstract}
% IC3/PDR is widely used for safety property checking in industrial RTL verification, but its performance depends on a web of interacting heuristics and implementation choices, making manual tuning costly and brittle.
% Motivated by this, we present \evolve, an offline automated framework that uses large language models to optimize IC3/PDR heuristics via slot-restricted, auditable code edits, retaining changes only when they pass proof-gated validation and compiling improvements into the checker with zero inference-time overhead.
% We use the public HWMCC suite as the optimization workload during evolution, and assess generalization and scalability on real industrial RTL benchmarks from production verification flows. The resulting evolved IC3/PDR engine achieves consistent performance improvements over strong baselines, establishing an initial and practical workflow for LLM-driven heuristic evolution in IC3/PDR.

IC3, also known as  property-directed reachability (PDR), is a commonly-used 
algorithm for hardware safety model checking. It checks if a state transition system complies with a given safety property. IC3 either returns UNSAFE (indicating property violation) with a counterexample trace, or SAFE with a checkable inductive invariant as the proof to safety. In practice, the performance of IC3 is dominated by a large web of interacting heuristics and implementation choices, making manual tuning costly, brittle, and hard to reproduce. 

This paper presents \evolve, an automated offline code-evolution framework that utilizes an LLM to propose small, slot-restricted and auditable patches to an IC3 implementation. Crucially, every candidate patch is admitted only through proof-/witness-gated validation: SAFE runs must emit a certificate that is independently checked, and UNSAFE runs must emit a replayable counterexample trace, preventing unsound edits from being deployed. Since the LLM is used only offline, the deployed artifact is a standalone evolved checker with zero ML/LLM inference overhead and no runtime model dependency.
We evolve on the public hardware model checking competition (HWMCC) benchmark and evaluate the generalizability on unseen public and industrial model checking benchmarks, showing that \evolve can reliably discover practical heuristic improvements under strict correctness gates.

%IC3/PDR (property-directed reachability) is a SAT-based, proof-producing algorithm for hardware safety model checking: it either returns a counterexample trace (UNSAFE) or a checkable inductive invariant (SAFE). In practice, however, its performance is dominated by a large web of interacting heuristics and implementation choices, making manual tuning costly, brittle, and hard to reproduce.
%We present \evolve, an automated offline code-evolution framework that uses an LLM to propose small, slot-restricted and auditable patches to an IC3/PDR implementation. Crucially, every candidate patch is admitted only through proof-/witness-gated validation: SAFE runs must emit a certificate that is independently checked, and UNSAFE runs must emit a replayable counterexample trace, preventing unsound edits from being deployed. Since the LLM is used only offline, the deployed artifact is a standalone evolved checker with zero ML/LLM inference overhead and no runtime model dependency.
%We evolve on public HWMCC instances and evaluate generalization on unseen public and industrial RTL benchmarks, showing that \evolve can reliably discover practical heuristic improvements under strict correctness gates.

\vspace{-10pt}
\end{abstract}

%% file: sections/intro.tex
\section{Introduction}
\label{sec:intro}
%\hongce{Overleaf comments are updated. Abstract is updated.}
% Formal property checking stands as a cornerstone of modern RTL verification. The escalating complexity of integrated circuits yields state spaces that far exceed what simulation can effectively explore, leaving deep corner-case behaviors under-tested and motivating formal, property-based analysis.

Formal property checking is a cornerstone of modern hardware verification. As integrated circuits grow in complexity, their state spaces quickly exceed what simulation can effectively explore, leaving deep corner-case behaviors untested, causing severe economic loss, thus motivating formal analysis in hardware verification, exemplified by hardware safety model checking.

% In practice, much of this workload is dominated by safety properties, ruling out reachable “bad” states, where performance and robustness directly determine how widely formal can be deployed. For safety checking, IC3/PDR~\cite{ic3,pdr} has emerged as a widely used, state-of-the-art algorithm in modern industrial model checkers, thanks to its SAT-driven, property-directed proof search and its ability to incrementally learn inductive invariants.
%In practice, much of this workload is dominated by safety properties, i.e., proving that no reachable execution can enter a ``bad'' state.

For hardware safety model checking, IC3~\cite{ic3} (also known as property-directed reachability, PDR~\cite{pdr}) is a widely-used state-of-the-art algorithm. Importantly, IC3 is \emph{proof-producing}: it either returns \texttt{UNSAFE} with a replayable counterexample trace, or returns \texttt{SAFE} with a checkable inductive invariant that certifies the property.

While IC3 provides a strong algorithmic foundation, its practical performance depends critically on numerous heuristics and engineering decisions---including generalization, clause propagation, and interactions with the underlying Boolean satisfiability (SAT) 
%(and sometimes SMT) 
engines. As a result, improving an IC3 checker is often more about heuristic engineering than changing the core algorithmic framework. Unfortunately, this engineering loop is costly and brittle~\cite{le_data_driven_ig,deepic3}: even small heuristic changes typically require re-running large benchmark suites to reach stable conclusions; gains on one family of instances can easily induce regressions elsewhere. Manual tuning therefore demands substantial domain expertise~\cite{le_data_driven_ig,deepic3}, making it difficult to reliably discover robust heuristic combinations.

\begin{figure}[t]
  \centering
  \includegraphics[width=\linewidth]{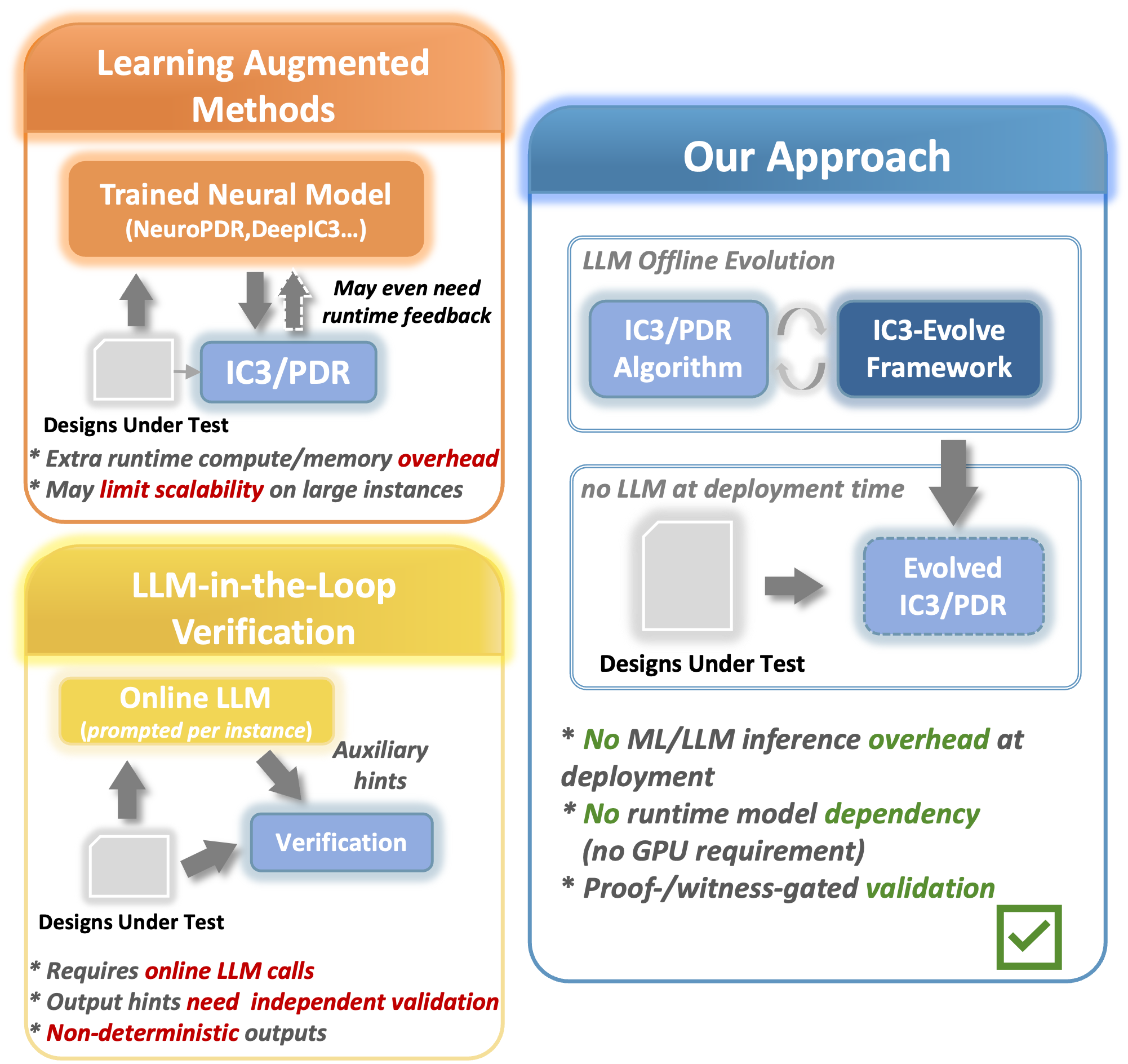}
  % \caption{Comparison of learning-augmented IC3, LLM-in-the-loop verification, and our approach. \evolve moves the LLM entirely offline to evolve IC3 code and admits edits only after proof-/witness-gated validation, producing a standalone evolved checker with zero model dependency at verification time.}
  \caption{Comparison of learning-augmented IC3, LLM-in-the-loop verification, and our approach. \evolve evolves IC3 code offline and promotes edits only via proof-/witness-gated validation, yielding a standalone checker with no runtime model dependency.}
  \label{fig:approaches_comparison}
  \vspace{-15pt}
\end{figure}

Recent advances in machine learning have also begun to influence Electronic Design Automation (EDA) and formal verification. In the context of IC3, prior work such as NeuroPDR~\cite{neuropdr} and DeepIC3~\cite{deepic3} explores learning-augmented solvers: a trained model predicts candidate clauses %(side-loaded lemmas) 
to guide IC3's proof search, showing promising results on selected benchmarks.

However, learning-augmented IC3 solvers often face practical barriers to industrial deployment. Learned components introduce training and inference costs, and their benefits can degrade on large-scale or heterogeneous workloads where robustness and throughput are paramount. Moreover, placing ML inference in the runtime loop adds latency and increases performance variance, prohibiting predictable behavior and integration into production flows. In recent HWMCC editions, the strongest-performing entries remain predominantly conventional model checkers rather than learning-augmented solvers~\cite{hwmcc24,hwmcc25}.

Another line of work uses LLMs \emph{in} the verification loop to provide per-instance auxiliary hints (e.g., candidate invariants or guidance) to an existing checker~\cite{janssen_chatgpt_verification,pirzada_llm_invariants_bmc,wu_llm_bmc_invariant}. Such hints must be independently validated before use; moreover, integrating online LLM calls at verification time raises practical concerns about latency, reproducibility, and deployment.

These trends suggest both an opportunity and a bottleneck: today's strongest IC3 checkers remain largely heuristic-engineered, yet manual tuning is labor-intensive and fragile. Meanwhile, recent LLM-based ``code evolution'' systems show that iterative propose--evaluate--refine loops can autonomously improve nontrivial algorithmic code under evaluation feedback. For example, AlphaEvolve~\cite{alphaevolve} demonstrates evaluator-driven evolutionary search for algorithmic optimization, while AutoSAT~\cite{autosat} and SATLUTION~\cite{satlution} apply related ideas to evolve SAT-solver heuristics under rigorous validation.

Motivated by this trend of research, we present \evolve, an offline LLM-driven framework that treats IC3 heuristic engineering as controlled code evolution. As shown in \textbf{\Cref{fig:approaches_comparison}}, rather than placing ML/LLM inference in the verification-time loop, \evolve uses LLMs to propose constrained, auditable code edits to an IC3 checker and promotes edits only after proof-/witness-gated validation. The resulted artifact is a standalone evolved checker: the LLM is used only offline, so deployment incurs no ML/LLM inference overhead and no runtime model dependency. Building on this foundation, we introduce a workflow that explores both local and cross-module heuristic choices and evaluate the evolved engine on public and industrial benchmarks.

We make the following contributions: \begin{itemize}
  \item We introduce \evolve, an offline LLM-driven framework for evolving IC3 heuristics through slot-restricted, auditable code edits, compiling validated improvements directly into the checker (with no inference-time overhead at deployment).

    % \item We design and implement a heuristic-evolution workflow that explores both local refinements and cross-module changes, including the \textbf{Compass \& Jump} strategy to guide search across the solver's modification space.

    \item We design and implement a heuristic-evolution workflow that explores local refinements and cross-module changes, including the \textbf{Compass \& Jump} strategy to guide search across the solver's modification space.

  \item We enforce correctness via proof-/witness-gated validation: a proposed edit is accepted only if it passes hard checks with validation, preventing unsound changes from being deployed.
% ---\texttt{SAFE} results must provide a checkable inductive invariant/certificate, while \texttt{UNSAFE} results must provide a replayable counterexample trace

  \item We evaluate \evolve by evolving on the public HWMCC suite and assessing generalization on unseen public and industrial verification benchmarks from production flows, demonstrating that \evolve can discover practical heuristic improvements under strict correctness gates.

\end{itemize}
% Upon acceptance, we will open-source the \evolve framework together with the evolved IC3 engine produced by our approach.
% Upon acceptance, we will open-source the \evolve and the evolved IC3 engine produced by our approach.

%% file: sections/preliminary.tex
\section{Background}\label{sec:background}

\subsection{IC3 in a Nutshell}
\label{subsec:ic3_nutshell}

IC3 is a proof-producing safety model checking algorithm: it tries to prove that no reachable execution can enter a ``bad'' state.
Operationally, it alternates between (i) searching for counterexamples to the current proof attempt and (ii) learning clauses that rule them out.
It either returns \texttt{UNSAFE} with a replayable counterexample trace, or returns \texttt{SAFE} with a checkable inductive invariant that certifies the property. Throughout the paper, we use proof to denote the SAFE certificate (inductive invariant), and witness to denote the UNSAFE counterexample trace.

We consider a finite-state transition system with state variables $\bm{x}$, input variables $\bm{y}$, initial-state predicate $I(\bm{x})$,
transition relation $T(\bm{x},\bm{y},\bm{x}')$, and a safety property $P(\bm{x})$, where $\bm{x}'$ denotes the next-state copy of $\bm{x}$.
IC3~\cite{ic3} decides whether all reachable states satisfy $P$.

IC3 maintains a sequence of \emph{frames} $F_0,\dots,F_k$, where $F_0 = I$ and each $F_i$ is represented as a formula over $\bm{x}$.
Intuitively, $F_i$ over-approximates the set of states reachable within $i$ steps.
The frames are monotone ($F_i \Rightarrow F_{i+1}$), and for $i\ge 1$ they are constrained to satisfy the property ($F_i \Rightarrow P$).
Additionally, the sequence satisfies the step condition: for all $0\le i<k$,
$F_i(\bm{x})\land T(\bm{x},\bm{y},\bm{x}') \Rightarrow F_{i+1}(\bm{x}')$.
IC3 strengthens this sequence by learning \emph{lemmas} that exclude unreachable states while preserving all reachable behaviors.\looseness=-1

% \textbf{\Cref{alg:ic3_overview}} sketches the main loop of IC3.
% At a high level, it alternates between two phases:
% (i)~\emph{blocking}, which finds a counterexample-to-induction (CTI) cube $s$ by checking
% $F_k \land T \land \neg P'$ (an $F_k$-state with a bad successor) for satisfiability and attempts to block $s$ by learning a new lemma; and
% (ii)~\emph{propagation}, which tries to push learned lemmas forward to higher frames.
% If blocking a proof obligation reaches level $0$, IC3 can construct a  counterexample trace and returns \texttt{UNSAFE}.
% If two adjacent frames become identical ($F_i = F_{i+1}$), that common frame is a 1-step inductive invariant implying $P$, and IC3 returns \texttt{SAFE}.

\textbf{\Cref{alg:ic3_overview}} sketches the main loop of IC3.
We use a \emph{cube} to denote a conjunction of literals representing a (partial) state assignment, and a \emph{clause} to denote
a disjunction of literals used as a lemma.
At a high level, IC3 alternates between two phases:
(i)~\emph{blocking}, which finds a counterexample-to-induction (CTI) cube $s$ by checking
$F_k \land T \land \neg P'$ (an $F_k$-state with a bad successor) for satisfiability and attempts to block $s$ by learning a new lemma; and
(ii)~\emph{propagation}, which tries to push learned lemmas forward to higher frames.
If blocking a proof obligation reaches level $0$, IC3 can construct a counterexample trace and returns \texttt{UNSAFE}.
If two adjacent frames become identical ($F_i = F_{i+1}$), that common frame is a 1-step inductive invariant implying $P$, and IC3 returns \texttt{SAFE}.\looseness=-1

\subsection{Key Subroutines and Heuristic Touchpoints}
\label{subsec:ic3_touchpoints}

\noindent\textbf{Key point.}
Most implementation choices in IC3 control \emph{search order} and \emph{resource allocation} (e.g., which obligation to tackle next, how aggressively to generalize, and how much SAT effort to spend).
They do not affect soundness as long as every learned lemma is validated by the corresponding SAT checks and \texttt{SAFE} is returned only at a validated fixpoint; under resource limits they mainly affect practical completeness via timeouts.
This makes these choices prime targets for systematic heuristic engineering (and for our offline code evolution later).

% \textbf{\Cref{alg:ic3_subroutines}} expands the key subroutines and makes explicit where heuristics dominate performance in practice.
% % We use a \emph{cube} to denote a conjunction of literals representing a (partial) state assignment, and a \emph{clause} to denote
% % a disjunction of literals used as a lemma.

\textbf{\Cref{alg:ic3_subroutines}} expands the key subroutines and makes explicit where heuristics dominate performance in practice.
We use the cube/clause terminology as defined above.
Most of IC3's runtime is spent answering incremental SAT queries of two types:
(1)~\emph{predecessor-search} queries (for recursive blocking), and
(2)~\emph{relative-inductiveness} queries (for generalization and pushing).

\paragraph{Proof obligations (\textsc{BlockProofObligations}).}
IC3 maintains a worklist $Q$ of proof obligations (POs) $(s,i)$, where $s$ is a cube to be blocked at frame $i$.
The policy for popping and re-inserting POs (e.g., depth-first vs.\ best-first, priority rules, aging) can strongly influence convergence:
it can bias the solver toward quickly refuting shallow CTIs (bug finding) or toward accumulating stronger lemmas for deeper proofs.
Practical implementations also rely on budgeting/backoff to avoid repeatedly spending SAT effort on ``stuck'' POs.
Since frames are continuously strengthened, queued POs can become stale and are typically skipped once they are already blocked by the current target frame (i.e., $F_i \land s$ becomes UNSAT).

\input{algorithms/IC3_algorithm_overview.tex}
\paragraph{Predecessor generalization (\textsc{PredGen}).}
When a PO $(s,i)$ cannot be blocked directly, IC3 queries $F_{i-1} \land T \land s'$ to obtain a predecessor cube $p$ and enqueues $(p,i-1)$.
\textsc{PredGen} attempts to generalize $p$ (e.g., by dropping literals or using partial models) while preserving reachability into $s$.
Stronger predecessor generalization can reduce search depth and the number of POs, but it typically requires extra SAT work and interacts heavily with the worklist strategy.

\paragraph{Inductive generalization (\textsc{IndGen}).}
Given a CTI cube $s$ at level $i$, \textsc{IndGen} derives a blocking clause $c \subseteq \neg s$ such that
$c$ is relatively inductive w.r.t.\ $F_{i-1}$ and does not exclude any initial state:
$I \Rightarrow c$ and $F_{i-1} \land c \land T \land \neg c'$ is UNSAT.
Most implementations start from $\neg s$ and attempt to drop literals while keeping relative inductiveness, yielding MIC-style generalization and its extensions.
They often leverage UNSAT cores/conflict clauses to guide minimization.
Optimizations here can substantially reduce SAT calls and improve lemma quality, but they introduce trade-offs between generalization strength and computational overhead~\cite{predictinglemmasgeneralizationic3}.

\paragraph{Push and propagation (\textsc{PushClauses}).}
To accelerate convergence, IC3 tries to push lemmas forward:
if a lemma $c \in F_i$ is also inductive relative to $F_i$, it can be added to $F_{i+1}$
(equivalently, if $F_i \land T \land \neg c'$ is UNSAT).
The effectiveness of pushing depends on clause ordering, caching, and simplification/subsumption policies, as well as on budgeting decisions that prevent spending too much effort on low-yield push attempts.

%% file: algorithms/IC3_algorithm_overview.tex
\begin{algorithm}[t]
\small
\caption{IC3 Algorithm Overview}
\label{alg:ic3_overview}
\SetAlgoLined
\SetKwInOut{Input}{Input}
\SetKwInOut{Output}{Output}

\SetKwFunction{BlockAll}{\algkw{kwBlock}{BlockProofObligations}}
\SetKwFunction{BlockOne}{\algkw{kwBlock}{BlockOne}}
\SetKwFunction{PushAll}{\algkw{kwPush}{PushClauses}}
\SetKwFunction{IndGen}{\algkw{kwIndG}{IndGen}}
\SetKwFunction{PredGen}{\algkw{kwPred}{PredGen}}

\SetKwProg{Fn}{Function}{:}{}
\BlankLine

\Input{$\langle I(\bm{x}),\, T(\bm{x},\bm{y},\bm{x}'),\, P(\bm{x})\rangle$}
\Output{\texttt{SAFE} or \texttt{UNSAFE}}
\tcp{$\bm{x}'$ denotes the next-state copy of $\bm{x}$; $P'$ abbreviates $P(\bm{x}')$}
\BlankLine

$F_0 \gets I$;\ $F_1 \gets P$;\ $k \gets 1$;\ $Q \gets \emptyset$\;
\tcp{$Q$ stores proof obligations $(s,i)$}
\If{$I \land \neg P$ is SAT or $I \land T \land \neg P'$ is SAT}{\KwRet{\texttt{UNSAFE}}}
\While{\texttt{true}}{
  \While{$F_k \land T \land \neg P'$ is SAT with witness cube $s$}{
    \If{\textbf{not} \BlockOne{$s,k$}}{\KwRet{\texttt{UNSAFE}}}
  }
  \PushAll{$k$}\;
  \If{$\exists i<k:\ F_i = F_{i+1}$}{\KwRet{\texttt{SAFE}}}
  $k \gets k+1$;\ $F_k \gets P$\;
}

\BlankLine
\Fn{\BlockOne{$s,k$}}{
  insert $(s,k)$ into $Q$\;
  \tcp{internally calls \PredGen and \IndGen while discharging $Q$}
  \Return \BlockAll{$Q$}\;
}
\end{algorithm}

% \begin{algorithm}[t]
% \small
% \caption{IC3 Key Subroutines}
% \label{alg:ic3_subroutines}
% \SetAlgoLined
% \SetKwProg{Fn}{Function}{:}{}
% \SetKwFunction{BlockAll}{\algkw{kwBlock}{BlockProofObligations}}
% \SetKwFunction{PushAll}{\algkw{kwPush}{PushClauses}}
% \SetKwFunction{IndGen}{\algkw{kwIndG}{IndGen}}
% \SetKwFunction{PredGen}{\algkw{kwPred}{PredGen}}
% \SetKwFunction{GetModel}{GetWitness}
% \BlankLine

% \Fn{\BlockAll{$Q$}}{
%   \While{$Q \neq \emptyset$}{
%     $(s,i) \gets \textsc{Pop}(Q)$\;
%     \lIf{$F_i\land s$ is UNSAT}{\textbf{continue}}
%     \If{$i=0$}{\Return \texttt{false}}
%     \If{$F_{i-1}\land T\land s'$ is SAT}{
%       $p \gets$ \PredGen{$\GetModel{},\, i-1,\, s$}\;
%       insert $(s,i)$ into $Q$\;
%       insert $(p,i-1)$ into $Q$\;
%     }
%     \Else{
%       $c \gets$ \IndGen{$s,i$}\;
%       \For{$j=1 \to i$}{ $F_j \gets F_j \land c$ }
%     }
%   }
%   \Return \texttt{true}\;
% }
% \BlankLine

% \Fn{\PredGen{$p, j, s$}}{
%   \tcp{Drop literals from predecessor cube $p$ while preserving reachability into $s$}
%   \Return $\textsc{MinimizePred}(F_j, T, p, s)$\;
% }
% \BlankLine

% \Fn{\IndGen{$s,i$}}{
%   \tcp{Return $c\subseteq\neg s$ such that $I\Rightarrow c$ and $F_{i-1}\land c \land T \Rightarrow c'$}
%   \Return $\textsc{RelInductiveMinimize}(I, F_{i-1}, T, \neg s)$\;
% }
% \BlankLine

% \Fn{\PushAll{$k$}}{
%   \For{$i=1 \to k-1$}{
%     \ForEach{$c \in F_i \setminus F_{i+1}$}{
%       \lIf{$F_i\land T\land \neg c'$ is UNSAT}{ $F_{i+1} \gets F_{i+1}\land c$ }
%     }
%   }
% }
% \end{algorithm}

\begin{algorithm}[t]
\small
\caption{IC3 Key Subroutines}
\label{alg:ic3_subroutines}
\SetAlgoLined
\SetKwProg{Fn}{Function}{:}{}
\SetKwFunction{BlockAll}{\algkw{kwBlock}{BlockProofObligations}}
\SetKwFunction{PushAll}{\algkw{kwPush}{PushClauses}}
\SetKwFunction{IndGen}{\algkw{kwIndG}{IndGen}}
\SetKwFunction{PredGen}{\algkw{kwPred}{PredGen}}
\SetKwFunction{GetModel}{GetWitness}
\BlankLine

\tcp{$s'$ and $c'$ denote the primed versions obtained by substituting $\bm{x}\mapsto \bm{x}'$}
\BlankLine

\Fn{\BlockAll{$Q$}}{
  \While{$Q \neq \emptyset$}{
    $(s,i) \gets \textsc{Pop}(Q)$\;
    \lIf{$F_i\land s$ is UNSAT}{\textbf{continue}}
    \If{$i=0$}{\Return \texttt{false}}
    \If{$F_{i-1}\land T\land s'$ is SAT}{
      $p \gets$ \PredGen{$\GetModel{},\, i-1,\, s$}\;
      insert $(s,i)$ into $Q$\;
      insert $(p,i-1)$ into $Q$\;
    }
    \Else{
      $c \gets$ \IndGen{$s,i$}\;
      \For{$j=1 \to i$}{ $F_j \gets F_j \land c$ }
    }
  }
  \Return \texttt{true}\;
}
\BlankLine

\Fn{\PredGen{$p, j, s$}}{
  \tcp{Drop literals from predecessor cube $p$ while preserving reachability into $s$}
  \Return $\textsc{MinimizePred}(F_j, T, p, s)$\;
}
\BlankLine

\Fn{\IndGen{$s,i$}}{
  \tcp{Return $c\subseteq\neg s$ such that $I\Rightarrow c$ and $F_{i-1}\land c \land T \Rightarrow c'$}
  \Return $\textsc{RelInductiveMinimize}(I, F_{i-1}, T, \neg s)$\;
}
\BlankLine

\Fn{\PushAll{$k$}}{
  \For{$i=1 \to k-1$}{
    \ForEach{$c \in F_i \setminus F_{i+1}$}{
      \lIf{$F_i\land T\land \neg c'$ is UNSAT}{ $F_{i+1} \gets F_{i+1}\land c$ }
    }
  }
}

\end{algorithm}

%% file: sections/method.tex
\section{Methodology}
\label{sec:method}
\subsection{\evolve Framework Overview}
\begin{figure*}[t]
  \centering
  \includegraphics[width=0.9\textwidth]{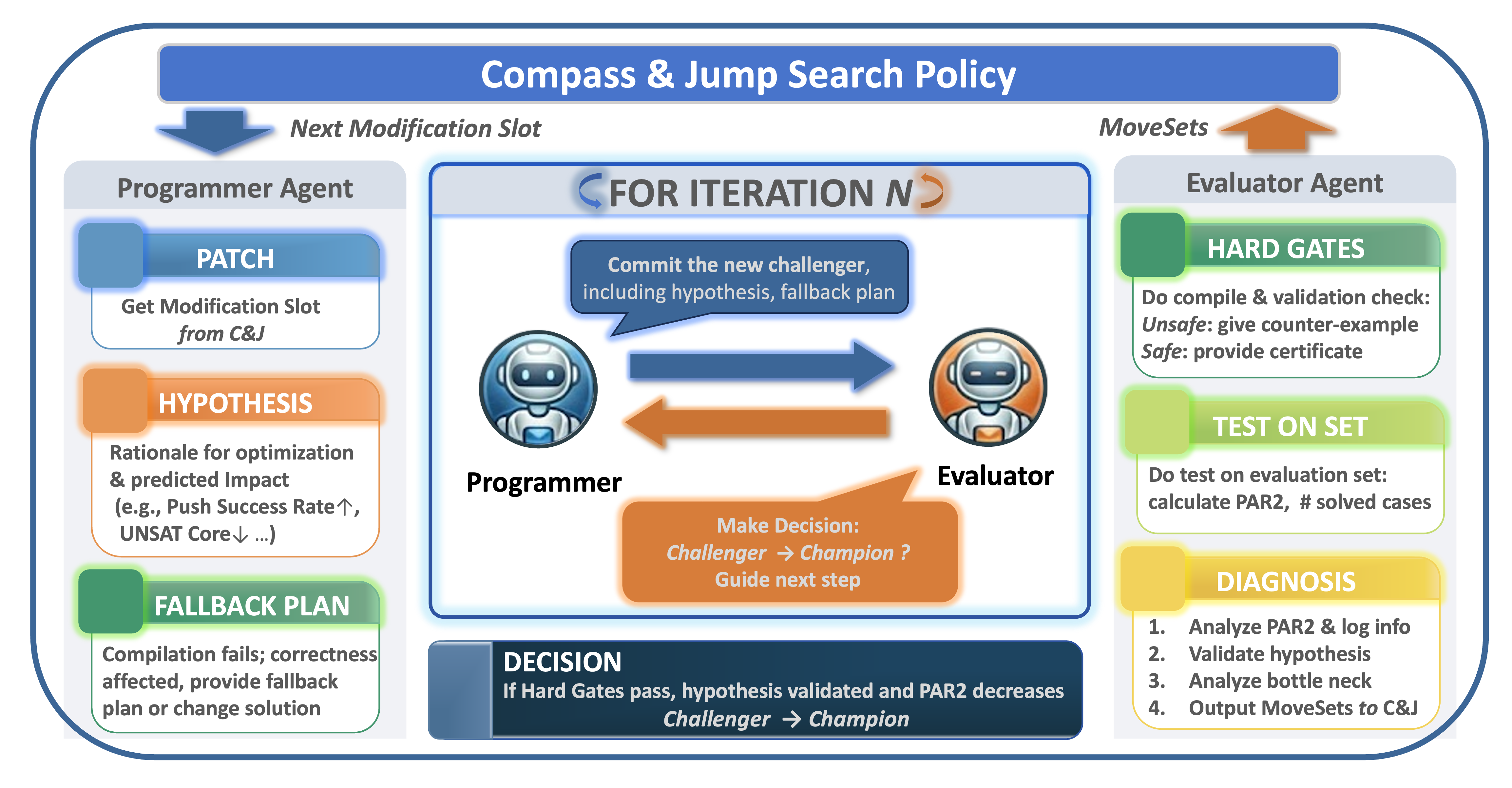}
    % \caption{Overview of \evolve. The Compass\&Jump search policy selects the next modification slot(s). A programmer agent proposes a slot-restricted patch together with a hypothesis and a fallback plan. An evaluator agent (tool-using) compiles the challenger and runs proof-/witness-gated validation as well as benchmarking (e.g., PAR2 and solved counts), and returns a diagnostic \emph{MoveSet} to guide the next iteration. A challenger is promoted to champion only if it passes hard gates and improves performance under a regression budget.}
    % \caption{Overview of \evolve. Compass\&Jump selects slot(s); a programmer proposes a slot-restricted patch (with hypothesis and fallback). A tool-using evaluator builds and witness-validates the challenger, benchmarks it (e.g., PAR2/solved), and returns a \emph{MoveSet}. Challengers are promoted only if they pass hard gates and improve under a regression budget.}
    \caption{Overview of \evolve. Compass\&Jump selects slot(s); a programmer proposes a slot-restricted patch (with hypothesis and fallback). A tool-using evaluator builds and witness-validates the challenger, benchmarks it (e.g., PAR2/solved), and returns a \emph{MoveSet}. Challengers are promoted only if they pass hard gates and improve under a regression budget. See agent prompt example in \Cref{app:prompts} in Appendix.}
  \label{fig:ic3_evolve_framework}
  \vspace{-12pt}
\end{figure*}

\textbf{\Cref{fig:ic3_evolve_framework}} summarizes \evolve, our offline framework for evolving IC3 implementations via iterative,
auditable code changes.
Starting from a \emph{champion} solver, each iteration selects a constrained modification scope (a set of heuristic slots) and asks
a programmer agent to propose a small patch together with a hypothesis about its expected impact and a fallback plan in case the
change proves invalid.
The candidate is then built and evaluated by an evaluator agent.
Crucially, we separate \emph{optimization} from \emph{correctness}: a patch is considered only if it passes proof-gated hard checks
with independent witness validation---\texttt{SAFE} runs must emit a checkable inductive invariant/certificate, and \texttt{UNSAFE}
runs must emit a counterexample trace that can be replayed by simulation.
Among the validated candidates, we retain improvements based on benchmark performance (e.g., PAR2 and solved counts) and use the
evaluator's diagnosis to steer subsequent iterations via the Compass\&Jump policy.

\subsection{Slot-Restricted Patch Space}
\label{subsec:slot_patch_space}
To make code evolution controllable and reviewer-auditable, our \evolve exposes a number of \emph{heuristic slots} that align with
IC3's dominant runtime subroutines (cf.~\Cref{subsec:ic3_touchpoints}).
Each iteration restricts the programmer agent to one (Compass) or a small set (Jump) of slots, so that proposed patches are
localized, interpretable, and easy to revert.
\textbf{\Cref{tab:heuristic_slots_compact}} summarizes the slots and representative optimization knobs.
Importantly, slots constrain \emph{where} we edit, not \emph{what} constitutes correctness: soundness is enforced exclusively by the
proof-gated validation described next.

\input{tables/slot_summary_compact.tex}

\paragraph{Slot-specific knowledge packs.}
A slot specifies a curated set of files/functions (and optionally line ranges) that the programmer agent is allowed to modify in a given iteration.
For each slot, we maintain a small, auditable knowledge pack that collects the most relevant external and internal references,
including solver code excerpts, open-source implementations, and slot-specific papers and documentation. 
At each iteration, the Compass\&Jump search policy retrieves and injects only the slot-relevant subset into the agents' prompts to reduce context noise and constrain edits toward well-understood patterns.
These knowledge packs only improve proposal quality and interpretability; correctness is enforced solely by the proof-/witness-gated validation.

\subsection{Two-Agent Closed-Loop Evolution}
\label{subsec:two_agent_closed_loop}
To operationalize heuristic engineering as a repeatable optimization process, \evolve runs a two-agent loop orchestrated by a lightweight Compass\&Jump search policy.
The framework maintains a current \emph{champion} implementation and applies the promotion rule (promote or revert) for each proposed challenger.
Validation and benchmarking are executed by the evaluator agent, which also produces a diagnostic \emph{MoveSet}---a ranked list of structured recommendations
$m=(\textsf{slot},\textsf{direction},\textsf{conf},\textsf{risk},\textsf{cost})$ summarizing the diagnosis and suggested next edits.
Each iteration records full provenance (patch diffs, logs, metrics, and witnesses) to keep the process auditable and to support post-hoc diagnosis.
\paragraph{Iteration protocol.}
Given a champion solver, one evolution iteration proceeds as follows:
\begin{enumerate}[leftmargin=*, label=\textbf{(\roman*)}, itemsep=0.15em, topsep=0.15em]
\item \textbf{Select scope.} The framework uses Compass\&Jump to choose the next modification slot(s) and assembles slot-relevant
      context (knowledge pack excerpts, code snippets, and the previous \emph{MoveSet} if available).
  \item \textbf{Propose.} The programmer agent produces a slot-restricted patch together with a hypothesis and a fallback plan.
\item \textbf{Validate.} The evaluator agent applies the patch, builds the challenger, and runs hard checks including  proof-/witness-gated validation. 

\input{algorithms/compass_jump_policy.tex}

% (\texttt{SAFE}: checkable invariant/certificate; \texttt{UNSAFE}: replayable counterexample trace).
\item \textbf{Test.} The evaluator agent evaluates the challenger on the evolution workload (a fixed HWMCC subset) under fixed timeouts,
      producing PAR2 and bucketed solved/timeout statistics.
\item \textbf{Diagnose \& guide.} The evaluator agent inspects the patch diff and evaluation artifacts, confirms (or refutes) the hypothesis, identifies bottlenecks, and returns a ranked MoveSet to guide the next slot and edit direction.\looseness=-1
\end{enumerate}
\noindent
The challenger is promoted only if it passes hard gates and improves the objective without unacceptable regressions; otherwise, the framework rolls back to the champion and continues with the next \emph{MoveSet}.

\subsection{Proof-/Witness-Gated Evaluation \& Promotion}
\label{subsec:proof_gated_eval_promotion}
\paragraph{Proof and witness hard gates.}
For each benchmark instance that a challenger solves within the timeout, we require it to emit an artifact that can be checked independently of the evolved solver.
If the solver reports \texttt{UNSAFE}, it must dump a counterexample trace that is replayed by \textsc{aigsim}~\cite{aiger_repo} to confirm that the trace indeed reaches a bad state.
If the solver reports \texttt{SAFE}, it must dump a proof (inductive invariant) that is validated by \textsc{certifaiger}~\cite{certifaiger}.
Any mismatch (missing proof/witness artifact, replay/check failure, or inconsistent return code) immediately rejects the challenger regardless of its performance.

\paragraph{Promotion and feedback on failures.}
Only challengers that pass the hard gates are scored on performance metrics such as PAR2 and solved counts.
When a challenger is rejected---either by witness validation or by failing to improve under the promotion rule---the evaluator inspects the diff and logs to produce a diagnosis and a ranked \emph{MoveSet}, which guides the programmer agent in the next iteration.
This proof-gated, witness-driven protocol ensures that evolution does not compromise the soundness of the underlying IC3 proof logic.

% \subsection{Search Policy: Compass\&Jump}
% \label{subsec:compass_jump}

% Compass\&Jump is a lightweight search policy that balances \emph{exploitation} and \emph{exploration} over the slot-restricted patch space.
% After each iteration, the evaluator agent returns a \emph{MoveSet}---a ranked list of candidate moves
% $m=(\textsf{slot},\textsf{direction},\textsf{conf},\textsf{risk},\textsf{cost})$.
% In \emph{Compass} mode, the policy takes a local step by selecting a single high-scoring move and restricting edits to its slot.
% In \emph{Jump} mode, the policy occasionally allows a small set of moves from \emph{distinct} slots (typically the jump size $J\in\{2,3\}$) to explore
% cross-slot synergies within one patch.

% To rank moves, we combine the evaluator-provided signals using a linear score
% $\textsf{score}(m)=w_c\cdot\textsf{conf}(m)-w_r\cdot\textsf{risk}(m)-w_k\cdot\textsf{cost}(m)$ with fixed weights $(w_c,w_r,w_k)$.
% The Jump probability $p_{\textsc{jump}}$ is adapted based on recent progress: it decreases when improvements are steady and increases under
% stagnation (details in \textbf{\Cref{alg:compass_jump_policy}}).
% Notably, Compass\&Jump search policy controls only search scope and guidance; correctness is guaranteed solely by the proof-/witness-gated validation in \textbf{\Cref{subsec:proof_gated_eval_promotion}}.
\subsection{Search Policy: Compass\&Jump}
\label{subsec:compass_jump}

Compass\&Jump is a lightweight \emph{search policy} for choosing the next editable slot(s) in the slot-restricted patch space.
After each iteration, the evaluator returns a \emph{MoveSet} $\mathcal{M}$, i.e., a ranked list of candidate moves
$m=(\textsf{slot},\textsf{direction},\textsf{conf},\textsf{risk},\textsf{cost})$.
We rank moves by a linear score
$\textsf{score}(m)=w_c\cdot\textsf{conf}(m)-w_r\cdot\textsf{risk}(m)-w_k\cdot\textsf{cost}(m)$ (fixed weights).
With probability $p_{\textsc{jump}}$ the policy enters \emph{Jump} mode and enables edits to the top $J\in\{2,3\}$ moves from \emph{distinct} slots,
encouraging cross-slot synergies; otherwise it enters \emph{Compass} mode and enables only the single best move (optionally preferring a lower-risk move when progress is volatile).
The jump probability $p_{\textsc{jump}}$ is adapted from recent progress (decrease under steady improvement; increase under stagnation).
Compass\&Jump controls only search scope and guidance; correctness is guaranteed solely by the proof-/witness-gated validation in \Cref{subsec:proof_gated_eval_promotion}.

%% file: tables/slot_summary_compact.tex
\begin{table}[t]
  \centering
  \footnotesize

  \setlength{\tabcolsep}{4pt}
  \renewcommand{\arraystretch}{1.1}
  \begin{tabularx}{\columnwidth}{@{}l >{\raggedright\arraybackslash}X@{}}
    \toprule
    \textbf{Slot} & \textbf{Scope (examples)} \\
    \midrule
    PO handling &
    \textsc{BlockProofObligations}: worklist order; stale-PO skip; requeue/budget \,/\, PO depth; SAT/PO; requeue rate \\
    \addlinespace
    Ind.\ gen. &
    \textsc{IndGen}: literal dropping; UNSAT cores; CTG/MIC budgets \,/\, core size; lemma yield; generalization SAT calls \\
    \addlinespace
    Pred.\ gen. &
    \textsc{PredGen}: assumption order; predecessor selection; cooldown \,/\, chain length; predecessor SAT calls; depth \\
    \addlinespace
    Push/prop. &
    \textsc{PushClauses}: push order; caching; subsumption \,/\, push success; redundancy; frame stabilization \\
    \addlinespace
    Cross-slot &
    Small multi-slot changes and synergies \,/\, overall PAR2 and bucketed regressions \\
    \bottomrule
    
  \end{tabularx}
  \caption{Heuristic slots for code evolution.}
  \label{tab:heuristic_slots_compact}
  \vspace{-10pt}
\end{table}

%% file: algorithms/compass_jump_policy.tex
\begin{algorithm}[t]
\small
\caption{Compass \& Jump search policy}
\label{alg:compass_jump_policy}
\SetAlgoLined
\SetKwInOut{Input}{Input}
\SetKwInOut{Output}{Output}

\Input{slots $\mathcal{S}$; MoveSet $\mathcal{M}$; progress history $\mathcal{H}$; jump prob.\ $p_{\textsc{jump}}$; jump size $J\in\{2,3\}$}
\Output{allowed slots $\mathcal{A}\subseteq\mathcal{S}$; guidance moves $\mathcal{G}$; updated $p_{\textsc{jump}}$}

\If{$\mathcal{M}=\emptyset$}{
  sample $s \sim \textsc{Uniform}(\mathcal{S})$\;
  \KwRet{$(\{s\},\, \emptyset,\, p_{\textsc{jump}})$}\;
}

$p_{\textsc{jump}} \gets \textsc{AdjustJump}(p_{\textsc{jump}}, \mathcal{H})$\tcp*{decrease if steady, increase if stagnant}
$\mathcal{M} \gets \textsc{RankByScore}(\mathcal{M})$\tcp*{using the linear score in \Cref{subsec:compass_jump}}

\If{$\textsc{Rand}() < p_{\textsc{jump}}$}{
  $\mathcal{G} \gets \textsc{TopDistinctSlots}(\mathcal{M}, J)$\;
}{
  $m^\star \gets \textsc{SelectBest}(\mathcal{M}, \mathcal{H})$\tcp*{optionally risk-averse when volatile}
  $\mathcal{G} \gets \{m^\star\}$\;
}
$\mathcal{A} \gets \{\, m.\textsf{slot} \mid m \in \mathcal{G}\,\}$\;
\KwRet{$(\mathcal{A},\, \mathcal{G},\, p_{\textsc{jump}})$}\;

\end{algorithm}

%% file: sections/experiment.tex
% todo: add some ref
\section{Experiments}\label{sec:experiment}

\subsection{Experimental Setup}
\label{subsec:exp_setup}
\paragraph{Platform.}
All experiments are conducted on Ubuntu 20.04.6 LTS, equipped with dual Intel\textsuperscript{\textregistered}
Xeon\textsuperscript{\textregistered} Platinum~8375C CPUs at 2.90\,GHz.

\paragraph{LLM model.}
We use GPT-5-Codex~\cite{codex} as the underlying model for both the programmer and evaluator agents during offline evolution.
% The model is never called at deployment time of model checkers.

\paragraph{Benchmarks.}
We evolve on a workload of 100 public HWMCC AIGER instances~\cite{hwmcc}, selected to cover a broad range of difficulty,
including trivial, non-trivial, and possibly timeout cases (spanning both \texttt{SAFE} and \texttt{UNSAFE}).
To assess transferability to production settings, we evaluate the final evolved engine on another 100 HWMCC instances (not used during evolution) and 302 industrial verification instances provided by industry.

\paragraph{Evolution configuration.}
We start from IC3ref~\cite{ic3ref_repo}, a lightweight textbook-style IC3 implementation with minimal built-in heuristics, which provides a clean and lightweight starting point for controlled evolution.
We run 200 offline evolution iterations in total.
In the first 100 iterations, we use a structured slot sweep to ensure broad coverage: edits are restricted to one slot at a time, and we move
to the next slot after several consecutive non-improving attempts.
In the remaining 100 iterations, we enable the Compass\&Jump search policy to select slots based on the evaluator’s \emph{MoveSet} feedback and
to occasionally allow multi-slot edits for exploring cross-slot synergies. Context-window management is described in \Cref{app:context_management} in Appendix.

\paragraph{Timeout and metrics.}
We use a timeout of 1800\,s  per instance and report PAR2 (the averaged solving time, where a timeout is penalized as taking $2\times$  the time limit). 
% We also report solved and timeout counts.

% \paragraph{Witness validation.}
% All reported \texttt{SAFE}/\texttt{UNSAFE} outcomes are independently validated as described in \textbf{\Cref{subsec:proof_gated_eval_promotion}}
% (\textsc{aigsim} for traces; \textsc{certifaiger} for certificates).

\subsection{Evolution Trace and Case Study}
\label{subsec:evolution_trace_case_study}
\begin{figure*}[t]
  \centering
  \includegraphics[width=0.95\textwidth]{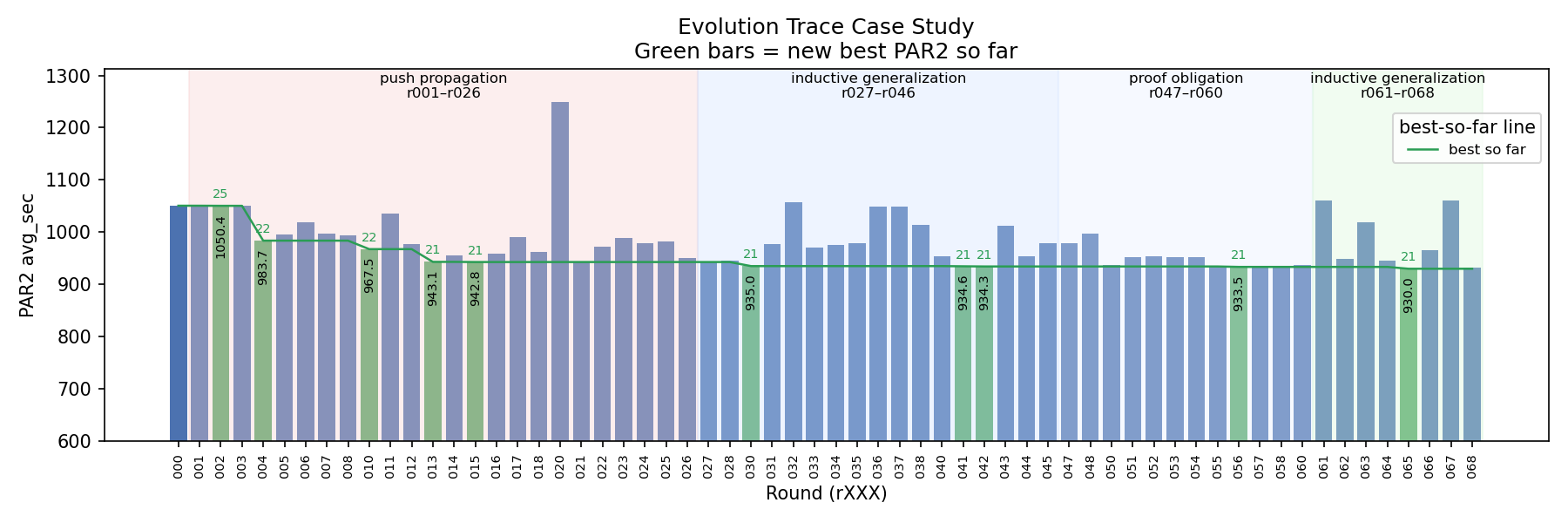}
  % \caption{A representative evolution trace on the HWMCC optimization workload.
  % Each bar corresponds to one attempted challenger; the line tracks the best-so-far PAR2 (lower is better).
  % Green bars indicate iterations that establish a new best-so-far solver variant, and shaded regions indicate which heuristic slot(s) were
  % exposed for editing during that phase.}
  \caption{Representative evolution trace on the HWMCC optimization workload. Bars: attempted challengers; line: best-so-far PAR2 (lower is better). Green bars: accepted new best; shading: editable slot(s) per phase.}
  \label{fig:evolve_iteration_example}
  % \vspace{-5pt}
\end{figure*}

\noindent
\textbf{\Cref{fig:evolve_iteration_example}} visualizes a typical offline evolution run.
PAR2 improvements arrive as a sequence of punctuated ``jumps'' in the best-so-far line, interleaved with many non-improving (and
thus rejected) challengers.
The shaded phases correspond to different exposed heuristic slots, illustrating how the evolution process explores and composes
improvements across solver subroutines while keeping correctness guarded by witness validation.

\paragraph{Case study: Push propagation (\textsc{PushClauses}).}
We zoom into rounds r000--r013, where edits are restricted to the \textsc{PushClauses} slot (all other code frozen).
Across three accepted edits, PAR2 decreases from 1050.61\,s to 943.07\,s and timeouts drop from 25 to 21.
Overall, the edits follow a consistent theme: avoid low-yield cleanup and reallocate effort toward frames with observed progress.

\noindent\textbf{Milestones (green bars in \textbf{\Cref{fig:evolve_iteration_example}}).}
We summarize best-so-far PAR2 (s) and timeouts (TO): \textbf{r000} 1050.61/25, \textbf{r002} 1050.42/25, \textbf{r004} 983.73/22, \textbf{r013} 943.07/21.

\noindent\textbf{r002: observe \& gate simplification.}
Simplification inside propagation can be expensive when few clauses are moving.
The edit adds lightweight progress observation and triggers simplification only after successful pushing or at periodic checkpoints:
{\footnotesize
\begin{verbatim}
if pushedThisRound or
   periodicCheckpoint():
    simplify()
record pushSuccessRate()
\end{verbatim}
}
This reduces low-yield cleanup without changing IC3's proof checks, and sets up later stall-aware scheduling.

\noindent\textbf{r004: skip repeatedly stalled frames.}
The next edit makes propagation stall-aware by tracking per-frame stall streaks and skipping push attempts on repeatedly unproductive frames:
{\footnotesize
\begin{verbatim}
if stallStreak[i] >= limit: continue
pushFrame(i)
updateStallStreak(i)  // reset
                      // else ++
\end{verbatim}
}
By redirecting effort to frames that still show improvement, the solver reduces churn and drops timeouts (25$\rightarrow$22) with a clear PAR2 gain.

\noindent\textbf{r013: stall-aware adaptive budgets.}
Finally, evolution learns an adaptive budgeting rule: allocate more push attempts to frames with recent successes, and cut budgets aggressively for deeply stalled frames with early termination:
{\footnotesize
\begin{verbatim}
budget <- adaptBudget(i)
// stall/success history
while clausesRemain(i) and budget>0:
    attemptPush(i)
    budget--
if earlyCut(i): break
\end{verbatim}
}
Compared to the binary skip/try rule, adaptive budgets continuously reallocate work based on observed yield, further improving PAR2 to 943.07\,s. A more detailed case-study interpretation is provided in \Cref{app:case_study_push} in Appendix.

\subsection{Comparison with Strong IC3 Baselines}
\label{subsec:baseline_comparison}
\textbf{\Cref{tab:baseline_comparison}} compares \evolve against several well-known IC3-based engines on public HWMCC benchmarks and an industrial benchmark suite. We report \#Solved (split into \#UNSAFE/\#SAFE) and PAR2 under the same timeout limit. The baselines include the vanilla IC3ref~\cite{ic3ref_repo}, IC3ref with a stronger SAT backend (IC3ref-CaDiCaL), ABC~\cite{abc} (whose IC3 engine is a widely-used industrial-strength model checker), and rIC3 as a strong recent IC3 implementation.

% A key fairness consideration is that a practical model checker is a pipeline that couples preprocessing, the model checking algorithm, and the SAT backend.
% Since \evolve targets \emph{IC3 heuristic/code evolution} rather than SAT-solver engineering, we control for the SAT backend by using CaDiCaL~\cite{cadical} in our evolved engine and comparing primarily against rIC3-CaDiCaL~\cite{gitsat-ric3-cadical}. This isolates improvements attributable to the IC3-level implementation choices.
% As shown in \Cref{tab:baseline_comparison}, \evolve achieves the best overall performance on both public sets (e.g., Set~B: 94 solved with PAR2 464.56) and transfers to the industrial suite (225/302 solved, PAR2 1044.26), outperforming all baselines.

A key fairness consideration is that a practical model checker is a pipeline that couples preprocessing, the model checking algorithm, and the SAT backend.
In particular, rIC3~\cite{ric3} is the recent HWMCC winner (HWMCC'24~\cite{hwmcc24} and HWMCC'25~\cite{hwmcc25}, in the bit-level safety tracks), making it a natural reference point for performance comparison.
Since \evolve targets \emph{IC3 heuristic/code evolution} rather than SAT-solver engineering, we control for the SAT backend by uniformly using CaDiCaL~\cite{cadical} in our evolved engine and comparing primarily against rIC3-CaDiCaL (rIC3 with CaDiCaL as the backend SAT solver, following the comparison setup in~\cite{gitsat-ric3-cadical}). This setup accurately measures improvements attributable to IC3-level implementation choices. \looseness=-1

\input{tables/comparison_with_baseline.tex}
\input{tables/one-slot-out-ablation.tex}

\subsection{Ablation Study}
% \paragraph{No-gate counterfactual (why proof-/witness gating is necessary).}
% To quantify the role of proof-/witness-gated validation, we report a counterfactual example from our evolution trace.
% Between rounds r093 and r094, the framework introduced a new \emph{clause-cleaning} heuristic, which would have increased the number of solved instances on the evolution workload by $+9$.
% A performance-only promotion rule would therefore accept this change.
% However, the challenger failed our proof-/witness gate (i.e., its reported \texttt{SAFE}/\texttt{UNSAFE} result could not be independently validated by certificate checking or trace replay) and was rejected.
% This illustrates why proof-/witness gating is essential in solver code evolution: it prevents unsound ``improvements'' from being promoted to champion and from propagating errors into subsequent rounds.\looseness=-1

\paragraph{No-gate counterfactual (why proof-/witness gating is necessary).}
In rounds r093--r094, \evolve proposed a new \emph{clause-cleaning} heuristic that would have increased solved instances on the evolution workload by $+9$, and would be promoted under performance-only selection.
However, the challenger failed independent witness validation (certificate checking / trace replay) and was rejected by our proof-/witness gate.
This counterfactual shows that witness gating is essential: it prevents unsound ``improvements'' from becoming champion and propagating into later rounds.\looseness=-1

\paragraph{Slot-isolated evolution (IC3 optimization is not a low-hanging fruit).}
We run four ablations where \evolve is restricted to a single heuristic slot for the entire evolution budget, meaning that all other subroutines remain frozen and only one ``knob family'' can be improved.
As shown in \textbf{\Cref{tab:ablation_slot_isolated}}, single-slot evolution yields only marginal and often unstable gains: improvements are small and may fail to transfer across sets (e.g., PO-only and Push-only improve little on Set~B and can even regress PAR2).
Moreover, naively composing the best single-slot edits (Naive-Compose) improves over the baseline but still falls far short of the full \evolve champion (e.g., Set~B: 85 vs.\ 94 solved, PAR2 775 vs.\ 465).
These results suggest that IC3 heuristics are tightly coupled: effective optimization requires coordinated, feedback-driven cross-slot evolution with regression-aware promotion, rather than isolated slot tweaking or \textit{ad hoc} ``pick-and-mix'' combinations.

%% file: tables/comparison_with_baseline.tex
\definecolor{rowhl}{RGB}{235,245,255}
\begin{table*}[t]
  \centering
  \small

  \vspace{-1em}
  \begin{threeparttable}

    \setlength{\tabcolsep}{3pt}
    \renewcommand{\arraystretch}{1.15}
    \begin{tabular}{@{}lcccccccccccc@{}}
      \toprule
      \multirow{2}{*}{\textbf{Solver}} &
      \multicolumn{4}{c}{\textbf{HWMCC Set A}\tnote{a}} &
      \multicolumn{4}{c}{\textbf{HWMCC Set B}\tnote{a}} &
      \multicolumn{4}{c}{\textbf{Industry Benchmark}} \\
      \cmidrule(lr){2-5} \cmidrule(lr){6-9} \cmidrule(lr){10-13}
      & \textbf{\#S/\#T}\tnote{b} & \textbf{\#UNSAFE} & \textbf{\#SAFE} & \textbf{PAR2(s)}
      & \textbf{\#S/\#T} & \textbf{\#UNSAFE} & \textbf{\#SAFE} & \textbf{PAR2(s)}
      & \textbf{\#S/\#T} & \textbf{\#UNSAFE} & \textbf{\#SAFE} & \textbf{PAR2(s)} \\
      \midrule
      IC3ref\tnote{c}       & 75/100 & 21 & 54 & 1050.61 & 75/100 & 26 & 49 & 1068.33 & 187/302 & 39 & 148 & 1403.31 \\
      IC3ref-CaDiCaL   & 85/100 & 22 & 63 & 718.87  & 85/100 & 28 & 57 & 732.28  & 198/302 & 38 & 160 & 1302.13 \\ \midrule
      ABC              & 86/100 & 22 & 64 & 665.59  & 88/100 & 26 & 62 & 618.59  & 185/302 & 36 & 149 & 1417.50  \\ \midrule
      rIC3-CaDiCaL     & 86/100 & 21 & 65 & 639.61  & 89/100 & 27 & 62 & 524.35  & 182/302 & 36 & 146 & 1449.78 \\
      \rowcolor{rowhl}
      \evolve          & \textbf{92}/100 & \textbf{24} & \textbf{68} & \textbf{490.67}
                       & \textbf{94}/100 & \textbf{28} & \textbf{66} & \textbf{464.56}
                       & \textbf{225}/302 & 36 & \textbf{189} & \textbf{1044.26} \\
      \bottomrule
    \end{tabular}
    \begin{tablenotes}[flushleft]
      \footnotesize
      \item[a] HWMCC Set A is the offline evolution set; Set B is unseen and not used in evolution..
      \item[b] \#S/\#T denotes \#Solved/\#Total (solved instances / total instances).
      \item[c] IC3ref serves as the starting point for our offline code evolution.
    \end{tablenotes}
  \end{threeparttable}
  \caption{Baseline comparison on public HWMCC and industrial benchmarks.}
  \label{tab:baseline_comparison}
  \vspace{-5pt}
\end{table*}

%% file: tables/one-slot-out-ablation.tex
\definecolor{rowhl}{RGB}{235,245,255}
\begin{table}[t]
  \centering
  \small
  \setlength{\tabcolsep}{4pt}
  % \caption{Slot-isolated evolution and naive composition ablation. ``PO-only/PredGen-only/IndGen-only/Push-only'' restrict \evolve to a
  % single slot for the entire evolution budget (all other code frozen). ``Naive-Compose'' merges the accepted edits from the four best
  % single-slot variants without further joint tuning. All results are computed under the same proof-/witness-gated protocol.}

  \vspace{-0.7em}
  
  \begin{tabular}{lcc}
    \toprule
    Variant &
    \shortstack{\textbf{HWMCC Set A}\\(\#Solved / PAR2(s))} &
    \shortstack{\textbf{HWMCC Set B}\\(\#Solved / PAR2(s))} \\
    \midrule
    Baseline                     & 75 / 1050.61  & 75 / 1068.33 \\
    \midrule
    PO-only               & 78 / 995.35 & 75 / 1093.42 \\
    PredGen-only          & 79 / 954.36 & 78 / 970.38 \\
    IndGen-only          & 78 / 927.80 & 84 / 803.73 \\
    Push-only             & 76 / 1016.05 & 75 / 1139.22 \\
    \midrule
    Naive-Compose  & 78 / 944.86 & 85 / 775.19 \\
    \midrule
     \rowcolor{rowhl}
    \evolve         & \textbf{92} / \textbf{490.67} & \textbf{94} / \textbf{464.56} \\
    \bottomrule
  \end{tabular}
    \caption{Slot-isolated and naive-composition ablation. Single-slot-only variants vs.\ Naive-Compose vs.\ full \evolve.}
    \label{tab:ablation_slot_isolated}
  \vspace{-10pt}
\end{table}

%% file: sections/conclusion.tex
\section{Conclusion and Future Work}
\label{sec:conclusion}
We presented \evolve, an offline LLM-driven code-evolution framework that proposes slot-restricted, auditable edits and admits them only through proof-/witness-gated validation, producing a standalone evolved checker with zero runtime model dependency.
Across public unseen/industrial workloads, \evolve delivers substantial performance gains, while our ablations indicate that improvements are not a low-hanging fruit: isolated single-slot evolution and naive composition fall far short of the full cross-slot champion.
More broadly, our results suggest that proof-producing verification offers a uniquely safe and reproducible setting for evaluator-driven LLM code evolution.
Future work will explore broader edit spaces beyond fixed slots, while maintaining auditability and regression control.

%% file: sections/discussion.tex
% Intentionally left blank.

%% file: sections/ack.tex
% Intentionally left blank.

%% file: algorithms/context_window_handle.tex
\begin{algorithm}[h]
\small
\caption{Prompt assembly with bounded context}
\label{alg:prompt_budget}
\SetAlgoLined
\KwIn{diff, metrics, logs, KB, slot focus}
\KwOut{prompt}
Assemble core fields (slot focus, diff summary, key metrics)\;
Attach local code snippets for target functions (budgeted)\;
Attach logs for top-$k\%$ slow cases (clip by lines/chars)\;
Attach slot-relevant KB snippets (cap length)\;
\If{prompt exceeds budget or agent output missing}{
  Drop KB; shorten logs/metrics/moveset; retry (slim prompt)\;
}
\end{algorithm}